# Exercising with an "Iron Man": Design for a Robot Exercise Coach for Persons with Dementia

**M Cooney, J Pihl, H Larsson, A Orand and E E Aksoy**

School of Information Technology, Halmstad University, 301 18 Halmstad, Sweden

martin.daniel.cooney@gmail.com

**Abstract.** Socially assistive robots are increasingly being designed to interact with humans in various therapeutical scenarios. We believe that one useful scenario is providing exercise coaching for Persons with Dementia (PWD), which involves unique challenges related to memory and communication. We present a design for a robot that can seek to help a PWD to conduct exercises by recognizing behaviors and providing feedback, in an online, multimodal, and engaging way. Additionally, in line with a mid-fidelity prototyping approach, we report on some feedback from an exploratory user study using a Baxter robot, which provided some confirmation of the usefulness of the general scenario; furthermore, the results suggested the degree of a robot's feedback could be tailored to provide impressions of attentiveness or fun. Limitations and possibilities for future improvement are outlined, touching on deep learning and haptic feedback, toward informing next designs.

## 1. Introduction

Dementia is a critical and growing problem. Although dementia cannot yet be cured, long-term therapist-led exercise at home could help to prevent deterioration of physical health, if the cost for human therapists could be reduced to a practical level [Hirsch 2013]. Here, as depicted in Fig. 1, we look toward socially assistive robots as potential exercise coaches, which have some additional benefits regarding supply, interest, reliability, and perception:

- Robots can be manufactured in required numbers and knowledge can be transferred computationally (there are also not enough human exercise therapy coaches).
- Robots can pique interest, as they are not yet common outside factories and laboratories.
- Robots can operate reliably by being also available outside of regular work hours; not becoming angry, sick, sweaty, or tired; and being sterilizable.
- (Some robots could be designed to sense signals related to exercising which humans cannot, such as heart rate or breathing.)

Such a robot would require the capability to deal with some unique challenges faced by persons with dementia (PWD). For example, the Alzheimer's Association and DSM-5 (Diagnostic and Statistical Manual of Mental Disorders) state that dementia involves impairments in areas such as memory and communication [A.A. 2019; A.P. 2013]. Memory is a vital part of learning; moreover, communication requires an ability to focus attention, recognize linguistic and visual social cues, and reason, as well as motivation--for example, PWD can find it difficult to conduct tasks when the pace is too fast or too slow, in accordance with the concepts of sensoristasis imbalance and the environment docility hypothesis [Kovach et al. 2000]. Impairments can also vary in degree; for the current paper we focus on the case of mild cognitive impairment, where we imagine any mistakes would be less critical and a human caregiver might not be required to help with the robot.

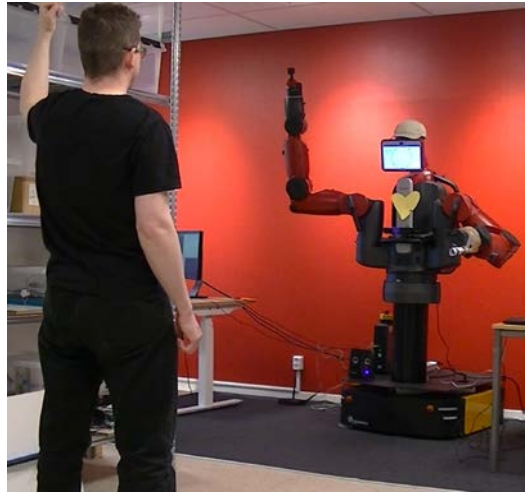

**Figure 1.** Basic concept: a robot can provide exercise coaching for Persons with Dementia (PWD)

Based on this, we propose that, for the context of robot exercise coaching for PWD, the following challenges should be considered:

- Memory: PWD can experience difficulty remembering how to do exercises or how many have been done.
- Communication: PWD can also experience trouble communicating needs and following a coach's instructions.

It is important to deal with these challenges because memory is required to follow protocols properly without doing too much or too little exercise, which could be damaging health-wise; furthermore, communication is required to alert in an emergency, and motivate in order to sustain regular exercise activities over time. We note that these categories, although not completely independent, are different: e.g., a PWD can be capable of communicating but not remember something, or remember a task but not be capable or willing to interact to carry it out.

Thus, the goal of the current paper was to gain insight into how to design a semi-autonomous robot exercise coach to guide PWD while addressing the challenges of impaired memory and communication. To do so, we employed a mid-fidelity prototyping approach based on grounded theory. First, a perception module was developed to analyze behaviors in an online and multimodal manner based on skeleton poses and gaze estimated from RGBD camera images, as well as speech; the output was subsequently used by a behavior module to seek to provide engaging, multimodal feedback. Finally, a simplified user study was conducted to explore how the system is perceived.

## 2. Related Work
Some previous work has explored how to design robots to care for PWD, or robots to be exercise coaches, but not robot exercise coaches for PWD.

*2.1. Robot Caregivers for PWD*

Various robots such as AIBO, NeCoRo, and Paro have been designed to offer perceived companionship [Yonemitsu et al. 2002, Libin et al. 2004, Wada et al. 2004]. Other robots including Pioneer, NAO, PaPeRo, Ryan, and ASCC offer useful services such as reminders, games, information about weather and time, and even cognitive evaluations [Tapus et al. 2009, Martin et al. 2013, Khosla et al. 2017, Abdollahi et al. 2017, Femandes et al. 2017]. Such robots address important needs but are not designed to guide a PWD in exercising.

*2.2. Robot Exercise Coaches*

There is a long history of robotics used to facilitate physical therapy and rehabilitation, from tool-like robot end effectors to exoskeletons [Chang et al. 2013]; recently some more human-like robots which could act like a therapist have also started to be be introduced. For example, one humanoid robot exercise coach was built by Matsusaka et al. with the intention to motivate elderly persons to exercise by sparking curiosity; the robot could be controlled by key presses or recognized speech utterances to demonstrate exercises alongside a human instructor [2009]. Some other work has also suggested how humans, including elderly, can exercise by walking with or chasing a robot [Hebesberger et al. 2017, Hansen et al. 2012].

Another work resembling our own is a study by Fasola and Mataric, who identified relevant design principles from psychology, described a multimodal implementation, and confirmed that user assessments of a robot were more positive than for a virtual agent [2013]. Although revealing much important information, this work was intended for users who lack cognitive impairments: From the prespective of memory, users were required to sit in front of a black curtain with a frontal face view and move their arms within a plane, but PWD might forget or not be able to follow these requirements; from the perspective of communication, users held a wireless controller to control the robot, which might not be possible for a PWD. Thus, for PWD, a design is required which takes into account such potential impairments in memory and communication.

*2.3. Human Exercise Therapy for PWD*

Programs such as the Seattle Protocols have been designed to help human caregivers and PWD to establish positive exercise experiences, by leveraging theories about social learning, gerontology, and dementia [Teri et al. 2008]. Such approaches comprise complex methods such as A-B-C (a form of cognitive behavior theory) which would not be practical for initial exploration, and are also not intended for robots, which can have different capabilities from humans (e.g. such as display screens).

Additionally, various general theories for behavior modification and devices exist: Flow Theory suggests that challenges should be matched with the person's skill level, an environment should be set up to allow a person to become absorbed (positive distraction), goals should be clear, and feedback should be rich and quick [Csikszentmihalyi 2013]. Models such as the integrative model of behavioral prediction, health belief model, social cognitive theory, and theory of reasoned action suggest that the importance of belief that the benefits of a desired activity outweigh costs [Fishbein et al. 2003]. Principles from Behavioral Economics can be applied by presenting a desired behavior as a default option, as an alternative to a less attractive option, or as something to do in the future [Lee et al. 2011]. And, Fogg's behavioral model proposes that, to elicit desired behavior, PWD should be motivated, have the ability to carry out the behavior, and perceive a trigger calling for the behavior [Fogg 2009]. Furthermore, a number of devices are being offered to facilitate memory and communication for PWD. For example, smart pill dispensers for medication, or easy-to-use phones with large buttons and reduced functionality. Such guidelines and products are informative but not sufficient to know how to design a robot exercise coach for PWD. Thus, dementia poses some unique challenges in relation to memory and communication, which the literature did not indicate how to handle for our context. The contribution of the current work is exploring a design for a robot exercise coach for PWD.

**3. Methods**
In this work we propose a design to address the highlighted challenges, then attempt to gain extra insight by implementing the design. We started by proposing three requirements for a robot exercise coach for PWD:

- (R1) Exercise Analysis: The robot should recognize if a person with dementia is exercising or not, what exercise they are doing, how many repetitions they have done, and the quality of the exercising, and use this information to provide quantitative and qualitative feedback.
- (R2) Interruption Detection: The robot should be able to control itself to a large extent, and adapt the flow of the interaction based on the perceived state of the PWD; i.e., it should recognize interruptions which could indicate a problem.
- (R3) Enjoyable Design: The robot should leverage knowledge from human science to make an enjoyable interaction for the PWD which motivates exercising.

R1 applies to memory, and R2 and R3 apply to communication (detecting a human's communications, and conveying the robot's communications, respectively). To address the requirements, we explored a design via prototyping with a robot:

- (A1) An online perception approach should be used. For our implementation we explored processing visual cues by reading 3D data from a Microsoft Kinect device.
- (A2) A multimodal approach should be used to flexibly detect interruptions and adapt. We explored identifying typical interruptions using visual and speech recognition.
- (A3) Engagement should be supported by leveraging motivational theory: Flow theory, integrative model of behavioral prediction, Fogg Motivation model, and Behavioral economics. For our implementation we explored providing quick fine feedback by having the robot mimick and score the person's movements.

A1-2 mainly concern the robot's perceptual module; A3 concerns the behavior module. Details are provided in the following sub-sections.

*3.1. Exercise Recognition*

To distinguish various kinds of exercises, we implemented a naive RGBD camera-based online exercise recognition module. The module tracks a person's arm movements using 3D skeleton pose data. To reduce the possibility of detecting false positives, a region of interest is defined relative to the user's body for each exercise. Based on the path the arm takes it is possible to infer whether or not the exercise was done correctly and counts as one repetition, or if it was done incorrectly and requires correction. This is done by calculating the length of an arm's path in the specified area and the angle between each vector that makes up the path using the following formulas:

$$length = \sum_{k=1}^{n-1} \sqrt{(p(k)_x - p(k+1)_x)^2 + (p(k)_y - p(k+1)_y)^2}$$

$$\theta = \arccos(\frac{\vec{p(k)p(k+1)} \cdot \vec{p(k+1)p(k+2)}}{|\vec{p(k)p(k+1)}| |\vec{p(k+1)p(k+2)}|})$$

(1)

where *n* is the number of points the path is made up of, *p* is the list of those points, and *k* is an index. The path length and the angles are then checked to be within some empirically determined limits; if they are, the exercise is deemed to have been done correctly and is then counted as a repetition. Depending on what was not within the limits the robot can then give feedback to the user on how to do it correctly.

A simplified check confirmed that the system accuracy seemed reasonable for our exploratory scenario: 88% for 60 exercises performed by three participants standing two meters from the robot. Accuracy was not perfect due to some fast motions not being detected, which we expect will be less problematic for PWD.

*3.2. Multimodal Attention-seeking*

To facilitate sustaining of attention, the system was set to conduct attention-grabbing behaviors when interruptions are detected. In addition to addressing a unique problem of PWD having short attention spans, such joint attention has also been described as important for successful social interactions, from the perspective of Joint Intention Theory and Situated Learning Theory [Breazeal et al. 2004]. The cause of interruptions can be internal (a person gets sleepy, bored, or agitated, and forgets what they were looking at) or external (something attention-grabbing in the environment, like an emergency such as a fire). As a first step, a simplified approach for speech recognition was implemented with CMU Pocketsphinx to detect some words related to controlling the robot or emergencies ("start", "stop", "help", "hurts", and "emergency"). As well, we considered that a patient who is paying attention would mostly have their head facing toward robot, with eyes open, an appropriate facial expression, and furthermore that they would be actively interacting and not saying anything about an emergency. Gaze could also be used to estimate what a person is looking at, but we used head pose which is easier to perceive, as a person's head is larger than their eyes and thus possibly easier to detect.

In detail, the system uses 2D-3D point correspondences between a detected face in an image and a generic model with camera parameters to calculate the pose of a person's head related to the robot's camera. To enhance processing speed, a region of interest is selected based on the location of the head in the pose detected by the Kinect. Contrast limited adaptive histogram equalization (CLAHE) is used to enhance the local contrast of the image to deal with varying illumination. Next, faces were found using Histogram of Oriented Gradients (HOG) and Support Vector Machines (SVMs), then facial landmarks were detected by refining an initial estimate using a cascade of regressors learned using gradient boosting [King et al. 2009]. Solving the Perspective-n-Point problem with a rough approximation of camera parameters and a generic 3d model of a face gives the rotation and translation vectors: an initial approximation from Direct Linear Transform (DLT) is iteratively refined to minimize the reprojection error using a combination of the Gauss–Newton algorithm (GNA) and gradient descent via Levenberg-Marquardt optimization. Also the eye aspect ratio (EAR) was calculated to estimate if a person's eyes are open [Soukupova et al. 2016]. Based on detected results, if a person is not paying attention for a set length of time, the robot was set to ask what the person was doing and wave its arm to seek to restore attention.

A simplified check was again conducted. Speech recognition in our context had an accuracy of 76% accuracy (65.3% with the robot's motors on and vibrating), where each word was repeated 15 times by three people one meter from the robot; we believe the accuracy could be substantially lower with PWD who might have problems with clearly enunciating in a loud voice, especially in a noisy environment. Head pose seemed reasonably robust at close distance, with 91.6% accuracy on 60 poses looking toward the robot or away (left or right).

*3.3. Engagement*

To facilitate engagement, the guidelines described in the related work section were incorporated into a simplified interactive scenario. At the start, users face a robot, which should offer a familiar communication interface to avoid giving stress, using multimodal humanoid channels such as facial expressions, voice, and gestures. Moreover, the robot should be safe to allow the users to concentrate, and could be large to be engaging (because tall robots are seen as dominant, conscientious, human-like, and emotionally stable, which are desirable qualities for a coach [Rae et al. 2013, Walters et al. 2009]). Based on this, we selected the Baxter robot as a platform, also designing behaviors intended to appear serious and caring, toward eliciting compliance [Goetz et al. 2002].

In adherence to cost-benefit analysis and Fogg's model, the robot briefly describes some of the benefits of exercising at the start of the interaction. Then exercises are conducted one by one. Flow Theory, cost-benefit analysis, and Fogg's model all suggest the importance of appropriate challenge, which was considered in selecting exercises to perform. Various forms of exercise have been prescribed for PWD from strength training, walking, gentle stretching, balance, and endurance. In this work, we focused on some common strength training exercises which can be done seated because many PWD have limited ability to move and stand for long periods of time, such exercises are simple (using one part of the body) and hence good for exploration, and challenge could be tailored by adjusting weights. Specifically, we focused on shoulder presses and side lateral raises. When starting an exercise, the robot clearly indicates the goal (what the user should do) and provides a trigger, stating when it is the user's turn. As the user exercises, feedback is provided through the robot's motions, speech, and display. During the interaction, distractions are minimized, to allow the user to concentrate, without providing excessive stimulation; for example, music was not used. Exercising was also set as the default option, as suggested in behavioral economics; the robot detects interruptions to end the interaction.

One question which arose was the *degree* to which feedback should be given--in other words, should a robot exercise coach mostly watch a person, or join in? To explore this question, we developed capability for our robot to mimic a user's arm movements during exercising, as shown in Fig. 2 and described below:

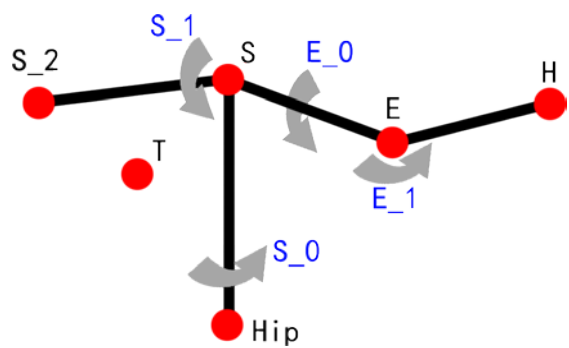

**Figure 2.** Model for mapping human to robot motions.

First, the Kinect coordinates of the human arm joints are mapped into the robot's coordinate space. Hip to shoulder and shoulder to hand distances are calculated and used to scale the human arm coordinates to fit the robot. Then angles were calculated using a model of the robot's joints. For example, when using the robot's left arm, normal vectors for the torso-plane, left shoulder and elbow were calculated.

$$\begin{aligned} \vec{T}_n &= \vec{TS}_2 \times \vec{TS} \\ \vec{down} &= \vec{T}_n \times \vec{S_2S} \\ \vec{S}_n &= \vec{down} \times \vec{ES} \\ \vec{E}_n &= \vec{HE} \times \vec{SE} \end{aligned} \qquad (2)$$

After this the relevant angles were calculated using dot products on the normalized vectors, and checked with thresholds to avoid tiny and extremely large movements due to instabilities in skeleton pose detection, before being sent to the robot.

$$\begin{aligned} S_0 &= -\arccos(\vec{T}_n \cdot \vec{S}_n) \\ S_1 &= \arccos(\vec{down} \cdot \vec{ES}) \\ E_0 &= -\arccos(\vec{T}_n \cdot \vec{E}_n) \\ E_1 &= \pi - \arccos(\vec{HE} \cdot \vec{SE}) \end{aligned} \qquad (3)$$

**4. Feedback**
We built a preliminary system based on concepts from the literature, but there was a question which arose in regard to how much feedback a robot should provide. To allow a PWD to concentrate and feel absorbed, minimal feedback could be desired which avoids over stimulation. Conversely, a turn-based approach might be more natural and familiar (conversations are usually turn-based), and therefore easy to understand. Or, the desirability of rapid fine feedback and control suggested that the robot could possibly continuously mimic the user's exercise motions. This has an analogy in the use of mirrors in rehabilitation (e.g., mirror therapy for stroke) and mirror neuron effect. Such mimicking could also be engaging because motion is a powerful visual saliency factor that exogenously drives attention. The robot can also provide company to the human and social priming by doing exercises together with a person, as if saying "Yes, it's hard, but I will do it with you." However, mimicking could also provide too much stimulus, and also compete with triggers in Fogg's model to conversely make it more difficult for users to become engaged.

*4.1. Participants*

To gain some basic exploratory insight into this question, we conducted a simplified within-subjects user study with 11 Swedish people at our university, who had never interacted with our robot before and received no monetary compensation. To also start to investigate effects of age and gender, around half of the participants were selected to be young adults (average age: 28.8 years, SD: 4.0) or elderly (average age: 69.0 years, SD: 11.5); also half were female, and half male.

*4.2. Procedure*

Each participant read a sheet of written instructions, provided consent, then interacted with three different versions of our system in random order. For each version of our system, the robot led the participant to perform two exercises, five times each. After interacting with a system, participants wrote down their overall impressions and filled out a questionnaire rating it. At the end, short interviews were conducted. Sessions took approximately 20 minutes.

*4.3. Conditions*

Participants each experienced the conditions below, in a random order.

System 1: Low-stimulus (no feedback).
System 2: Turn-based (feedback without mimicking).
System 3: Continuous (feedback with mimicking).

*4.4. Measures*

Participants wrote down their overall impressions of the system, and also filled out a questionnaire containing the following questions:
Q1 Memory: The robot helped me to keep track of the number of repetitions I had done.
Q2 Memory: The robot helped me to do the exercises correctly.
Q3 Communication: The robot could handle interruptions.
Q4 Communication: The robot made the exercise session engaging.
Q5 Overall: I liked interacting with the robot.

The study was carried out in a grounded theory-style fashion, with the objective of gaining basic insight, since we did not know ahead of time how the robot would be perceived.

*4.5. Results*

Typical impressions of our system voiced by more than one participant are shown in Table 1. A few participants who conducted exercises quickly described sometimes observing some delays, which could be confusing; such effects have been described elsewhere (e.g. SpeechJammer) [Kurihara et al. 2012], although this might not be such a large problem for PWD, who we expect will perform exercises more slowly than our participants.

**Table 3.** Impressions from participants for each system. The number in parentheses is the number of participants who reported an impression.

| System | Impressions |
| --- | --- |
| 1 No feedback | Annoying (6) |
| 2 Turn-based | Helpful (10), Attentive (8) |
| 3 Mimicking | Helpful (10), Confusing (3), Fun (2) |

Questionnaire results are shown in Fig.3. A three-way repeated measures analysis of variance (ANOVA) was conducted on the questionnaire scores with factors age, gender, and robot design (1-3).

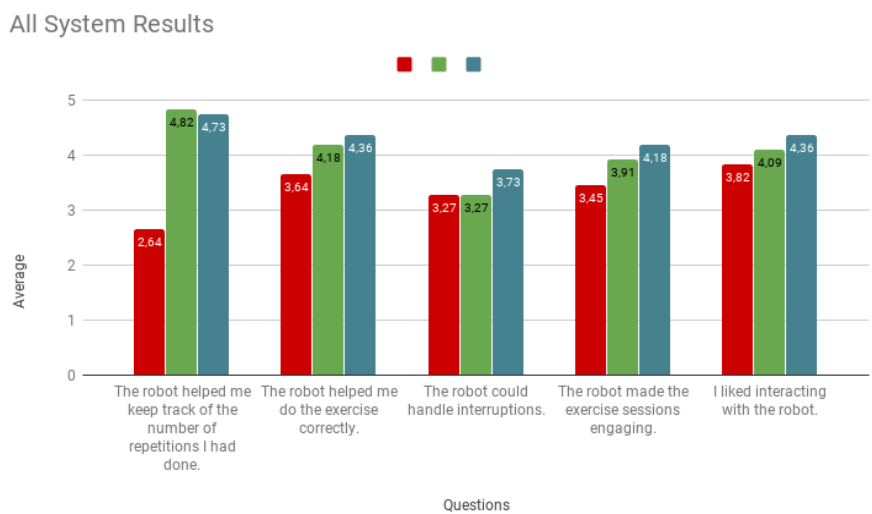

**Figure 3.** Questionnaire results regarding degree of feedback.

For Q1 about the robot helping repetition counting, there was no significant interaction effect. An ANOVA with a Greenhouse-Geisser correction indicated that perception of the systems differed ($F(1.153, 8.072) = 14.046$, $p = .005$, $\eta^2 = .667$). Post-hoc analysis with the Bonferroni adjustment indicated that participants found system 1 to be less helpful than systems 2 and 3 ($p = .011$ and $p=.029$). For Q2 about the robot helping to do exercises correctly, there was a statistically significant interaction between the effects of age and system on participants' perception, $F(2, 14) = 5.057$, $p = .022$, $\eta^2 = .419$. Simple effects analysis indicated that young participants felt that system 1 provided significantly less help than systems 2 and 3: $p = .015$ and $.002$ respectively. For Q3 about handling interruptions, there was a statistically significant interaction between the effects of age, gender, and system on participants' perception that the robot could handle interruptions, $F(2, 14) = 4.763$, $p = .026$, $\eta^2 = .405$. Simple effects analysis indicated that older male participants thought that system 3 was better than systems 1 and 2, possibly because they could see quickly when the third system stopped moving that it had detected an interruption.

For Q4 about engagement, interaction effects of age and system were observed: $F(2, 14) = 4.096$, $p = .040$, $\eta^2 = .369$. Young participants found system 1 to be significantly less engaging than system 2 and 3 ($p = .015$, $p=.028$). For Q5 about preferences, participants liked interacting with all versions of the robot to some degree (one-sample t-tests compared with a neutral value of 3 yielded $p =.03$, $p=.006$, $p=.001$); a significant difference between versions was not observed due to variances in preference, suggesting the usefulness of further work to detect potential correlates such as Big 5 Factors or anxiety toward robots, and use this to personalize feedback levels in robot exercise coaches.

## 5. Discussion

In summary, the contribution of the current work is in exploring an initial design for a semi-autonomous robot exercise coach which can lead exercises for PWD in an online, flexible and engaging manner based on recognizing people's exercise behaviors and providing feedback. This work is summarized in an online video at https://youtu.be/GqEOJZB5YeM.

In detail, from a number of known problems which affect PWD, we proposed two which we believe to be particularly relevant to the context of exercising: memory and communication. Next we presented a mid-fidelity solution comprising a perceptual and behavioral module based on RGBD data and sound, which was used to gain insight in a simplified grounded theory-style check. The results suggested that exercising with a robot coach can be enjoyable, that feedback is considered helpful, and that less feedback can enhance perceived attentiveness and more feedback can be perceived as fun.

*5.1. Limitation and Future Work*

The main limitation of the current work is its exploratory nature: simplified approaches were used to gain some exploratory insight into a new scenario of using robots as exercise coaches for PWD. Future work will include conducting a user study with PWD and more participants, as well as investigating more powerful recognition approaches involving deep learning methods, and alternative feedback modalities: Our current proposed perception module is naive and relies on heuristics. We are currently implementing a convolutional long short-term memory (convLSTM) based neural network architecture which takes RGBD image sequences and returns detected exercises while computing the 3D pose information in between. In the same manner, we are planning to extend the network architecture with more branches to additionally detect the attention. The result will be a compact perception framework working in an end-to-end manner.

Additionally we are investigating integrating our system with a haptic feedback system developed in previous work focused on bilateral training for hemiplegics [Orand et al. 2019], which could offer engaging and rich feedback, although questions rise regarding implementation and how to structure feedback. A wearable design would offer simplicity at the cost of possibly requiring assistance to attach the sensors to a PWD; conversely, a more complex solution could involve the robot touching the PWD (e.g., using its own hand to push a person's arm to the left or right to correct an exercise). Moreover, should haptic feedback be used when a repetition is completed, or when an arm's trajectory deviates, and if so, how? Our aim is that exploring such questions could bring us a step closer to an age of more accessible therapy--through socially assistive robots.

**Acknowledgments**
We thank all those who kindly contributed help and thoughts!